%% file: main.tex
\pdfoutput=1

\documentclass[10pt,twocolumn,letterpaper]{article}

\usepackage[pagenumbers]{cvpr}   %

\input{preamble}

\definecolor{cvprblue}{rgb}{0.21,0.49,0.74}
\usepackage[pagebackref,breaklinks,colorlinks,allcolors=cvprblue]{hyperref}

\def\paperID{*****} %
\def\confName{CVPR}
\def\confYear{2026}

\title{GraftRay: Physically Based Mesh Insertion into Gaussian Splatting Scenes \\ via Path Tracing}

\author{Jinxin Ai\\
Technical University of Munich\\
{\tt\small jinxin.ai@tum.de}
\and
 Matthias Niessner\\
 Technical University of Munich\\
 {\tt\small niessner@tum.de}
}

\begin{document}

\twocolumn[{%
\renewcommand\twocolumn[1][]{#1}%
  \maketitle
  \begin{center}
    \includegraphics[width=\textwidth]{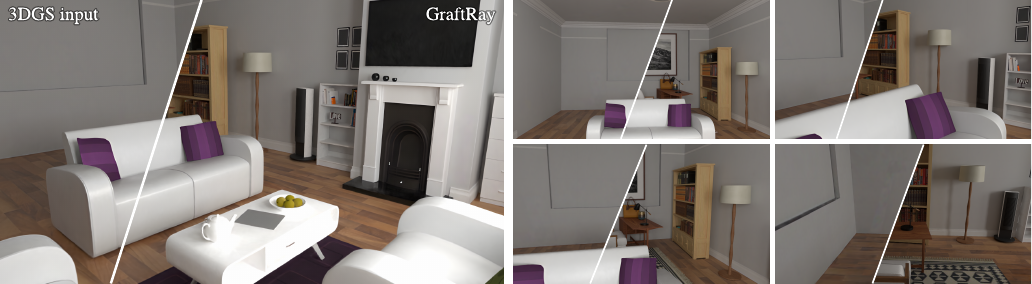}
    \captionof{figure}{\textbf{GraftRay} inserts mesh objects into a 3D Gaussian Splatting scene with physically consistent shading and shadows.}
    \label{fig:teaser}
  \end{center}%
}]

\input{sec/0_abstract}
\input{sec/1_intro}
\input{sec/2_related}
\input{sec/3_method}
\input{sec/4_experiments}
\input{sec/5_conclusion}

{
    \small
    \clearpage
    \bibliographystyle{ieeenat_fullname}
    \bibliography{main}
}

\end{document}

%% file: preamble.tex
\usepackage{multirow}
\usepackage{tabularx}

\newcommand{\method}{GraftRay\xspace}

\newcommand{\bx}{\mathbf{x}}
\newcommand{\by}{\mathbf{y}}

\newcommand{\bn}{\mathbf{n}}

\newcommand{\bmu}{\boldsymbol{\mu}}
\newcommand{\bL}{\mathbf{L}}
\newcommand{\bI}{\mathbf{I}}
\newcommand{\bB}{\mathbf{B}}
\newcommand{\ba}{\mathbf{a}}

\DeclareUnicodeCharacter{2194}{\ensuremath{\leftrightarrow}}

\makeatletter
\newlength\gT\newlength\gX\newlength\gcw\newlength\grh\newlength\ggw\newlength\glw
\newsavebox\g@rowbox\newsavebox\g@imgbox
\newcommand\gfont{\small}
\newcommand\gsetup[5]{\setlength\gT{#1}\def\g@tw{#2}\def\g@th{#3}\def\g@gap{#4}\def\g@W{#5}}
\newcommand\g@dims{\setlength\gX{\dimexpr\gT-\glw\relax}%
  \setlength\gcw{\dimexpr\gX*\g@tw/\g@W\relax}\setlength\grh{\dimexpr\gX*\g@th/\g@W\relax}%
  \setlength\ggw{\dimexpr\gX*\g@gap/\g@W\relax}}
\newsavebox\g@labbox
\newcommand\g@rlab[2]{\sbox\g@labbox{\gfont #2}%
  \ifdim\wd\g@labbox<\dimexpr#1+1pt\relax\rotatebox{90}{\makebox[#1]{\usebox\g@labbox}}%
  \else\rotatebox{90}{\parbox[c]{#1}{\centering\gfont #2}}\fi}
\newcommand\g@rows[1]{\vbox{\offinterlineskip\def\g@sep{}\@for\g@i:=#1\do{\g@sep
  \hbox{\g@rlab{\grh}{\g@i}}\def\g@sep{\vskip\ggw}}}}
\newcommand\g@head[1]{\hbox{\def\g@sep{}\@for\g@i:=#1\do{\g@sep\makebox[\gcw]{\gfont\strut\g@i}\def\g@sep{\hskip\ggw}}}}
\newcommand\ggrid[3]{\leavevmode\begingroup\def\g@r{#1}\def\g@c{#2}%
  \ifx\g@r\@empty\setlength\glw{0pt}\g@dims\else
    \setlength\glw{12pt}\g@dims\sbox\g@rowbox{\g@rows{#1}}%
    \setlength\glw{\dimexpr\wd\g@rowbox+1.5pt\relax}\g@dims\sbox\g@rowbox{\g@rows{#1}}\fi
  \vbox{\offinterlineskip
    \ifx\g@c\@empty\else\hbox{\hskip\glw\g@head{#2}}\vskip1pt\fi
    \hbox{\ifx\g@r\@empty\else\hbox to\glw{\usebox\g@rowbox\hss}\fi\includegraphics[width=\gX]{#3}}}\endgroup}
\newcommand\gfixed[1]{\setlength\glw{#1}\g@dims}
\newcommand\ghead[1]{\hbox{\hskip\glw\g@head{#1}}}
\newcommand\gstrip[3][]{\sbox\g@imgbox{\includegraphics[width=\gX]{#3}}%
  \hbox{\hbox to\glw{\hss\ifx\relax#1\relax\raisebox{\dimexpr\ht\g@imgbox/2-0.25em\relax}{\parbox[c]{\glw}{\raggedright\gfont #2}}%
    \else\g@rlab{\ht\g@imgbox}{#2}\fi\hss}\usebox\g@imgbox}}
\makeatother

%% file: sec/0_abstract.tex
\begin{abstract}
Inserting virtual objects into scenes reconstructed via 2D or 3D Gaussian Splatting (2DGS/3DGS) is a key capability for content creation, AR, and synthetic data generation. However, realistic insertion requires the object to be seamlessly lit by the scene and, conversely, to cast physically plausible shadows onto it. Existing approaches either harmonize appearance in image space, sacrificing multi-view consistency, or approximate scene illumination with an environment map at the insertion point, which cannot represent near-field area lights such as windows and lamps that dominate indoor lighting. To address this, we present GraftRay, a unified method for physically consistent insertion of mesh objects into indoor scenes represented by either 2D or 3D Gaussians. We recover scene materials and surface normals with diffusion-based priors, and explicitly model spatially-extended emitters, solving their radiance in closed form. A hybrid renderer then jointly path-traces the scene Gaussians and inserted meshes, producing view-consistent shading, interreflections, and soft shadows. On FIPT, GraftRay sets a new state of the art, outperforming recent baselines such as ComGS and EAG-PT in rendering fidelity and shadow accuracy, while maintaining highly competitive rendering efficiency.
\end{abstract}

%% file: sec/1_intro.tex
\section{Introduction}
\label{sec:intro}

3D Gaussian Splatting (3DGS)~\cite{kerbl20233d} and its 2D counterpart~\cite{huang20242d} enable high-quality, real-time rendering of reconstructed scenes. However, inserting a virtual object into such scenes remains challenging. Beyond matching its appearance to the background, the inserted object must respond to the scene's illumination, cast geometrically consistent shadows, and contribute indirect light to surrounding surfaces. Achieving these effects requires explicit light transport while preserving the visual fidelity of the original reconstruction.

\noindent Existing approaches address this problem from three main directions. \emph{Environment-map-based methods} estimate local illumination from HDR environment maps~\cite{bolduc2025gaslight,liang2024photorealistic,huang2026lightharmony3d}. Although effective for approximating incident illumination, environment maps do not explicitly represent the spatial positions of local emitters or their interactions with scene geometry, limiting their ability to reproduce spatially varying illumination and geometrically consistent shadows. \emph{Inverse-rendering methods} jointly recover scene materials and illumination~\cite{liang2024gs,du2025gs,gu2025irgs,zhang2023neilf++,lin2025iris,sabae2026aegir,chen2024gi}. While physically grounded, these methods introduce additional unknowns beyond those strictly required for object insertion. Moreover, low-dynamic-range (LDR) observations weaken the photometric constraints needed to distinguish emissive sources from highly reflective surfaces. Finally, \emph{image-space harmonization methods} use generative models to refine the appearance of composited objects~\cite{ren2026mv,violante2026lighting,skorokhodov2025diffusion,li2025freeinsert}. Despite producing visually plausible results, they do not explicitly model scene-level light transport and therefore cannot guarantee consistent relighting when the object or illumination changes.

\noindent We propose \textbf{GraftRay}, a unified framework for physically consistent object insertion into 2D and 3D Gaussian scenes (\cref{fig:teaser}). Because our pipeline extracts material priors directly from rendered views rather than relying on original capture data, it uniquely supports physically-based insertion even for synthetically generated Gaussian scenes where original training images are unavailable. We first render multi-view images from the Gaussian scene and estimate material properties, including albedo, roughness, metallicity, and normals, using robust 2D estimators such as RGB-X or diffusion-based models. These predictions are then unprojected onto individual 2D/3D Gaussians to provide scene-level material attributes, bypassing the complex joint optimization of traditional inverse rendering. Building on these attributes, we extend an OptiX-accelerated path tracer to support Gaussian primitives with stochastic transparency~\cite{enderton2010stochastic} and $3\sigma$ axis-aligned bounding boxes, traced jointly with inserted meshes. The extended path tracer incorporates a Mitsuba-style path integration scheme~\cite{Mitsuba3}, analytic area emitters, multiple importance sampling (MIS), and explicit shadow rays to enable light transport directly on Gaussian scenes. We further recover scene illumination from LDR observations by treating saturated pixels as lower-bound constraints and using unsaturated surfaces to constrain emitter intensities. Since light transport is linear in emitter intensities with fixed geometry and materials, we precompute unit-radiance basis images and formulate light estimation as a nonnegative least-squares (NNLS) problem. Finally, we use differential rendering~\cite{debevec2008rendering} to introduce the inserted object's shadows and indirect illumination while preserving the original background reconstruction.

\noindent Our contributions are summarized as follows:
\begin{itemize}
    \item From LDR images alone, we unproject multi-view 2D material and normal predictions onto individual 2D/3D Gaussians, providing scene-level material attributes for physically based insertion without joint inverse rendering.

    \item We extend an OptiX path tracer to trace stochastic-transparency Gaussians jointly with meshes, computing illumination, shadows, and indirect light between Gaussian scenes and mesh objects, while differential rendering preserves the reconstruction.

    \item We formulate light-intensity estimation using censored observations and path-traced radiance bases, exploiting unsaturated illumination cues to distinguish true emitters from bright reflectors and efficiently estimate nonnegative emitter intensities.
\end{itemize}

%% file: sec/2_related.tex
\begin{figure*}[t]
  \centering
  \includegraphics[width=\linewidth]{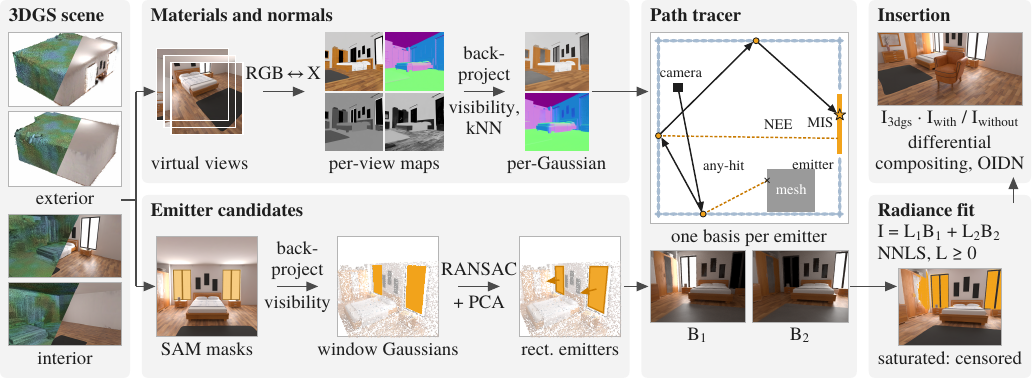}
  \caption{\textbf{Overview of \method.} From LDR photographs we reconstruct a
  3DGS scene (tiles: Gaussians as feature-coloured ellipsoids $|$ render) and lift
  materials and normals onto it (\cref{sec:prelim}). SAM-marked lights become
  rectangles (\cref{sec:candidates}); a Gaussian path tracer (\cref{sec:tracer})
  renders one basis image per emitter, whose radiances are solved by censored least
  squares (\cref{sec:radiance}). Objects are inserted by differential
  compositing (\cref{sec:composite}).}
  \label{fig:pipeline}
\end{figure*}

\section{Related Work}
\label{sec:related}

\paragraph{Lighting estimation for object insertion.}
Classical object insertion relies on differential rendering~\cite{debevec2008rendering}, which combines a captured scene with an estimate of its illumination; Karsch \etal~\cite{karsch2011rendering} extended this paradigm to single photographs. Subsequent methods estimate illumination from images using learned HDR environment maps~\cite{gardner2017learning}, spatially varying lighting~\cite{garon2019fast,li2020inverse}, or lighting volumes~\cite{srinivasan2020lighthouse}. Diffusion-based approaches further improve illumination estimation and adaptation: DiffusionLight~\cite{phongthawee2024diffusionlight} synthesizes environment-map observations through a rendered chrome ball, while DiPIR~\cite{liang2024photorealistic} uses a diffusion model as a critic for lighting optimization. GaSLight~\cite{bolduc2025gaslight} lifts ordinary images into HDR Gaussian splats, LightHarmony3D~\cite{huang2026lightharmony3d} predicts environment maps for hybrid Gaussian--mesh rendering, and LuxRemix~\cite{liang2026luxremix} decomposes indoor illumination into editable components. Unlike approaches that primarily estimate environment maps or distribute emissive radiance across scene representations, GraftRay explicitly parameterizes local area emitters and uses them in path-traced light transport, allowing their spatial extent and scene occlusion to affect illumination and shadows.

\paragraph{Compositing and harmonization in 3D.}
Recent methods use generative priors to improve the visual consistency of inserted objects in reconstructed scenes. MV-CoLight~\cite{ren2026mv} generates multi-view-consistent lighting and shadows, while Violante \etal~\cite{violante2026lighting} harmonize composite images with diffusion models and consolidate the results into 3DGS. D3DR~\cite{skorokhodov2025diffusion} similarly uses diffusion priors for zero-shot 3DGS harmonization. Other approaches explore alternative scene representations and insertion strategies: ComGS~\cite{gao2026comgs} composes objects using surface octahedral probes, and FreeInsert~\cite{li2025freeinsert} inserts text-specified objects without requiring explicit spatial priors. Related 2D methods include image harmonization~\cite{cong2020dovenet}, diffusion-based object relighting~\cite{neuralgaffer}, token-space relighting from sparse views~\cite{chen2026lumitokens}, and geometry-guided video insertion with GO-Renderer~\cite{gu2026go}. GaussianEditor~\cite{chen2024gaussianeditor} explores direct editing of 3DGS scenes. These methods primarily target perceptual consistency or generative scene editing, whereas GraftRay explicitly models light transport to produce view-consistent shadows and indirect illumination for inserted objects.
\paragraph{Inverse rendering on Gaussian splats.}
Neural inverse rendering initially developed around radiance fields~\cite{mildenhall2020nerf}, factorizing geometry, reflectance, and illumination~\cite{zhang2021nerfactor,jin2023tensoir} with ray tracing for indoor scenes~\cite{zhu20232}. For Gaussians, GS-IR~\cite{liang2024gs} and GS-ID~\cite{du2025gs} recover per-Gaussian BRDFs under environment maps or spherical Gaussians. GI-GS~\cite{chen2024gi} and IRGS~\cite{gu2025irgs} incorporate global illumination, with IRGS employing differentiable 2D Gaussian ray tracing. NeILF++~\cite{zhang2023neilf++} couples incident and outgoing light fields, while IRIS~\cite{lin2025iris} addresses indoor LDR reconstruction by estimating camera responses. Intrinsic diffusion priors~\cite{zeng2024rgb,kocsis2024intrinsic} and factorized inverse path tracing~\cite{wu2023factorized} offer alternative material decomposition strategies. 
Among these, AEGIR~\cite{sabae2026aegir} is closely related, jointly optimizing area emitters, materials, and geometry via a differentiable deferred renderer. Similarly, EAG-PT~\cite{yang2026eag} reconstructs emission-aware 2DGS with albedo only for path-traced editing. Unlike these joint factorizations, GraftRay targets object insertion with a constrained optimization: we initialize scene materials from priors and optimize emitter radiances while keeping geometry and materials fixed, reducing coupled unknowns and enabling explicit handling of saturated LDR observations.

\paragraph{Ray tracing Gaussian primitives.}
Recent methods have extended ray tracing to Gaussian and particle-based representations. 3DGRT~\cite{moenne20243d} traces particle scenes using ray-tracing hardware, bounding individual primitives and evaluating their responses along rays. EVER~\cite{mai2025ever} supports exact ellipsoid rendering for view synthesis, while GHPT~\cite{bo2026ghpt} combines rasterization and path tracing for real-time relightable Gaussian splats. Building on these efforts, we extend an OptiX-accelerated path tracer to support light transport directly on Gaussian scenes. Our approach shares the peak-response intersection strategy with 3DGRT but targets physical illumination rather than view synthesis. Stochastic hit acceptance \cite{sun2025stochastic} accounts for Gaussian transparency during ray traversal, enabling Monte Carlo light-transport estimation with analytic area emitters, multiple importance sampling, and next-event estimation.

%% file: sec/3_method.tex
\section{Method}
\label{sec:method}

Given a reconstructed Gaussian Splatting scene and
user-marked light regions, our method estimates scene illumination
and inserts objects with consistent direct and indirect lighting.
As illustrated in \cref{fig:pipeline}, we first estimate per-Gaussian
materials and normals from multi-view intrinsic-image predictions.
We then initialize analytic area-light candidates from the marked
regions and estimate their radiances using a Gaussian-aware path
tracer and unit-radiance basis renders. Finally, differential
compositing transfers the lighting changes induced by the inserted
objects to the original 3DGS rendering while preserving its
view-dependent appearance.

\subsection{Material and Normal Estimation}
\label{sec:prelim}

A Gaussian Splatting scene consists of $N$ anisotropic Gaussians with means
$\bmu_i$, rotations $R_i$, scales
$S_i=\operatorname{diag}(\mathbf{s}_i)$, opacities $o_i$, and
view-dependent colours~\cite{kerbl20233d}. Physically based light
transport additionally requires a surface normal $\bn_i$ and
BRDF parameters, including albedo $\ba_i$, roughness $\rho_i$,
and metallic $m_i$, which are not explicitly provided by the
reconstruction. We estimate these attributes by rendering the scene
from $V$ virtual views and applying the intrinsic-image diffusion
model RGB$\leftrightarrow$X~\cite{zeng2024rgb} with $S$ random seeds
to each view. The predicted material and normal maps are then
back-projected onto the Gaussians.

\paragraph{Visibility.}
To avoid assigning a pixel's prediction to occluded Gaussians,
we determine the visible surface depth from accumulated
transmittance along each camera ray. Specifically, we define
the surface depth $z$ as the depth at which the transmittance
first falls below $\tfrac{1}{2}$, corresponding to the median
depth of the accumulated opacity. A Gaussian at camera-space
depth $r$ is accepted only if
\[
\frac{|r-z|}{z}<\tau_z,
\]
where $\tau_z$ is a depth tolerance. This test rejects Gaussians
that lie in front of or behind the visible surface.

\paragraph{Aggregation.}
For each Gaussian, we first average predictions from all
accepting views separately for each random seed, and then take
the per-channel median across seeds. This aggregation reduces
stochastic variation in the diffusion predictions and
disagreements across views. Predicted normals are transformed
to world space before aggregation and renormalized afterwards.
We keep the resulting materials and normals fixed during
illumination estimation, thereby avoiding joint optimization
of lighting and material parameters.

This per-Gaussian back-projection turns predictions that are
plausible per image but mutually inconsistent into a single
scene-level estimate. A 2D estimator may assign the same
surface a different albedo or roughness in different views or
seeds; since each Gaussian is visible in many views, the
average over its accepting views acts as a multi-view
consensus that suppresses view-specific errors, and the median
over seeds removes outlying samples. Storing the result on the
Gaussians also makes materials available wherever a path
arrives, including surfaces outside the current view.

 \subsection{Emitter Initialization}
\label{sec:candidates}

Each emitter is manually specified by the user through a 2D mask, while its geometry and radiance are estimated automatically. The user marks light-emitting regions, such as windows, lamps, and ceiling panels, using SAM~\cite{kirillov2023segment} in one or a few views. We back-project the masked pixels onto the Gaussians using the visibility test in \cref{sec:prelim}, obtaining a 3D point set for each emitter. Each emitter is represented as an analytic, one-sided rectangle $E_j$, parameterized by its centre, in-plane axes, half-extents, and normal. We fit a plane to each point set using RANSAC weighted towards bright Gaussians, then refine the plane using PCA on the inliers. This weighting favours bright surfaces when estimating the emitter geometry. For wall-mounted emitters, we align the in-plane axes with the horizontal and vertical directions; otherwise, we use the PCA axes. The emitter's extent is determined from the Gaussians associated with the fitted plane, and its normal is oriented towards the room. The resulting emitter candidates are subsequently refined during radiance estimation in \cref{sec:radiance}, allowing candidates with little contribution to the observed illumination to be suppressed.
\subsection{Path Tracing on Gaussian Scenes}
\label{sec:tracer}

\paragraph{Gaussians as custom primitives.}
We represent each Gaussian by an axis-aligned bounding box
enclosing its $3\sigma$ ellipsoid. For a ray
$\bx(t)=\mathbf{o}+t\mathbf{d}$, we transform the ray into
the Gaussian's local frame:
\begin{equation}
\mathbf{p}
=
S_i^{-1}R_i^{\!\top}(\mathbf{o}-\bmu_i),
\qquad
\mathbf{q}
=
S_i^{-1}R_i^{\!\top}\mathbf{d}.
\end{equation}
The Gaussian response along the ray reaches its maximum at
\begin{equation}
t_i^\star
=
-\frac{\mathbf{p}\cdot\mathbf{q}}{\|\mathbf{q}\|^2},
\qquad
m_i^2
=
\|\mathbf{p}\|^2
-
\frac{(\mathbf{p}\cdot\mathbf{q})^2}
{\|\mathbf{q}\|^2},
\label{eq:gaussray}
\end{equation}
where $m_i^2$ is the squared Mahalanobis distance at the
closest point on the ray line. The intersection program reports
$t_i^\star$ when $m_i^2<9$ and assigns the effective opacity
\begin{equation}
\alpha_i=o_i\exp\left(-\frac{1}{2}m_i^2\right).
\end{equation}
The reported intersection must also lie within the valid
ray interval.

\paragraph{Stochastic transparency.}
Alpha compositing requires intersections to be processed in
depth order, whereas hardware ray traversal visits primitives
in an arbitrary order. Following stochastic
transparency~\cite{enderton2010stochastic}, the any-hit program
accepts an intersection with Gaussian $i$ with probability
$\alpha_i$. OptiX then returns the nearest accepted intersection.
Consequently, the probability that Gaussian $i$ is the first
accepted primitive is
\begin{equation}
w_i
=
\Pr(K=i)
=
\alpha_i
\prod_{j\in\mathcal{F}_i}(1-\alpha_j),
\label{eq:hit-prob}
\end{equation}
where $\mathcal{F}_i$ denotes the primitives in front of $i$.
This probability equals the corresponding alpha-compositing
weight. Thus, shading the returned primitive provides a
single-sample estimate of its contribution to the composited
radiance. The complete estimator must also account for any
background contribution when the ray misses all accepted
primitives.
Compared with collecting and sorting all intersections along
a ray, this requires no per-ray intersection buffer: each path
segment is a single hardware traversal, and the inserted
meshes are traced in the same acceleration structure. The random acceptance is
averaged out by the samples per pixel like the other Monte
Carlo dimensions.

\paragraph{Shadow rays.}
Shadow rays use a separate ray type that evaluates
transmittance deterministically rather than sampling
stochastic intersections. Its any-hit program accumulates
transmittance by multiplying it by $(1-\alpha_i)$ and
terminates when the value falls below $5\times10^{-3}$.
This continuous transmittance estimate supports smooth
soft shadows at the sample counts used for basis rendering.

\subsection{Light Transport}
\label{sec:transport}

For each pixel, we average radiance estimates over jittered
camera rays sampled across the pixel footprint. Along each
ray, the outgoing radiance is composited according to
\begin{equation}
L_{\mathrm{ray}}
=
\sum_i w_i L_{\mathrm{o}}(\bx_i,\mathbf{v}),
\label{eq:ray-blend}
\end{equation}
where $w_i$ is defined in \cref{eq:hit-prob} and
$L_{\mathrm{o}}$ is evaluated using the rendering equation
with a Cook--Torrance BRDF and GGX microfacet model. Emitters are represented as analytic rectangles rather than
Gaussians. We estimate direct illumination using next-event
estimation on the rectangles and combine it with BSDF sampling
through multiple importance sampling with the power heuristic
~\cite{10.1145/218380.218498,pharr2016physically}.
Indirect illumination is estimated by continuing paths along
BSDF-sampled directions, with Russian roulette used for
path termination. Our integrator follows Mitsuba's
\texttt{path} integrator~\cite{Mitsuba3}, replacing mesh
intersections with the Gaussian primitives described above.

\begin{table*}[!tp]
\centering
\setlength{\tabcolsep}{2.6pt}
\resizebox{\linewidth}{!}{%
\begin{tabular}{@{}l ccccccc c ccccccc c cc c c@{}}
\toprule
& \multicolumn{15}{c}{Synthetic (path-traced ground truth)} && \multicolumn{2}{c}{Real} && Generated\\
\cmidrule{2-16}\cmidrule{18-19}\cmidrule{21-21}
& \multicolumn{7}{c}{FIPT} && \multicolumn{7}{c}{NFINITE} && FIPT-Real & ScanNet++$^\ast$ && SpAItial$^\ast$\\
\cmidrule{2-8}\cmidrule{10-16}
Method & $\Delta$PSNR$\uparrow$ & Mask$\uparrow$ & IoU$\uparrow$ & RMSE$\downarrow$ & PSNR$\uparrow$ & SSIM$\uparrow$ & LPIPS$\downarrow$ && $\Delta$PSNR$\uparrow$ & Mask$\uparrow$ & IoU$\uparrow$ & RMSE$\downarrow$ & PSNR$\uparrow$ & SSIM$\uparrow$ & LPIPS$\downarrow$ && VQA ratio$\uparrow$ & VQA ratio$\uparrow$ && VQA ratio$\uparrow$\\
\midrule
DOT3D~\cite{violante2026lighting} & 19.33 & 12.27 & 0.18 & 0.39 & 17.40 & 0.785 & 0.409 && 20.75 & 11.31 & 0.16 & 0.33 & 20.45 & 0.861 & 0.217 && \textbf{0.401} & \textbf{0.443} && 0.464\\
MV-CoLight~\cite{ren2026mv} & 25.58 & 20.27 & 0.46 & 0.27 & 22.90 & 0.824 & 0.451 && 26.53 & 18.16 & 0.38 & 0.19 & 25.27 & 0.829 & 0.332 && 0.392 & 0.440 && 0.509\\
GS-IR~\cite{liang2024gs} & 26.71 & 21.18 & 0.47 & 0.18 & 23.81 & 0.853 & 0.387 && 29.10 & 21.49 & 0.45 & \textbf{0.17} & 27.14 & 0.893 & 0.210 && 0.385 & 0.412 && 0.478\\
GI-GS~\cite{chen2024gi} & 26.62 & 21.48 & 0.50 & 0.17 & 24.16 & 0.854 & 0.387 && 28.19 & 21.43 & 0.38 & 0.20 & 26.85 & 0.892 & 0.210 && 0.371 & 0.416 && 0.512\\
GaSLight~\cite{bolduc2025gaslight} & 18.36 & 12.85 & 0.13 & 0.35 & 17.58 & 0.763 & 0.424 && 17.43 & 10.50 & 0.15 & 0.32 & 19.68 & 0.857 & 0.256 && 0.369 & 0.421 && 0.494\\
IRGS~\cite{gu2025irgs} & 28.40 & 22.56 & 0.50 & 0.15 & 25.16 & 0.868 & 0.380 && 29.50 & 22.06 & \textbf{0.52} & \textbf{0.17} & 27.67 & 0.897 & 0.206 && 0.373 & 0.413 && 0.504\\
EAG-PT~\cite{yang2026eag} & 24.26 & 21.19 & 0.40 & 0.29 & 21.67 & 0.770 & 0.519 && 15.89 & 12.16 & 0.13 & 0.49 & 13.70 & 0.642 & 0.447 && 0.355 & -- && --\\
ComGS~\cite{gao2026comgs} & 13.09 & 8.27 & 0.26 & 0.33 & 13.93 & 0.721 & 0.454 && 14.86 & 7.76 & 0.22 & 0.39 & 16.48 & 0.827 & 0.250 && 0.334 & 0.365 && 0.435\\
LumiTokens~\cite{chen2026lumitokens} & 21.27 & 15.98 & 0.00 & 0.40 & 19.49 & 0.787 & 0.429 && 24.22 & 15.43 & 0.00 & 0.36 & 23.16 & 0.866 & 0.212 && 0.354 & 0.395 && 0.464\\
GO-Renderer~\cite{gu2026go} & 15.89 & 10.58 & 0.02 & 0.62 & 15.29 & 0.694 & 0.497 && 21.56 & 13.58 & 0.01 & 0.59 & 20.42 & 0.788 & 0.293 && 0.395 & 0.415$^\ast$ && 0.488\\
\midrule
Ours (DiffusionRenderer) & \textbf{35.17} & \textbf{29.14} & 0.76 & 0.08 & \textbf{29.81} & \textbf{0.895} & \textbf{0.350} && 31.30 & 23.91 & 0.51 & 0.18 & 28.21 & 0.902 & 0.182 && 0.369 & 0.442 && 0.516\\
Ours (RGB$\leftrightarrow$X) & 34.29 & 28.49 & \textbf{0.77} & \textbf{0.07} & 29.27 & \textbf{0.895} & 0.351 && \textbf{32.51} & \textbf{25.11} & 0.49 & \textbf{0.17} & \textbf{28.66} & \textbf{0.904} & \textbf{0.180} && 0.368 & 0.441 && \textbf{0.538}\\
\bottomrule
\end{tabular}}
\caption{\textbf{Object insertion.} All methods use the same objects, lights and
three views per scene. Synthetic scenes are scored against path-traced ground truth
($\Delta$PSNR, Mask, IoU and RMSE measure the insertion effect); real and generated
scenes, which have none, by the no-reference VQAScore ratio. GS-IR, GI-GS, IRGS: their
recovered light in our renderer. --: not run. $^\ast$Three of the twelve real and generated
scenes (ScanNet++ 8b5caf3398, a24858e51e; Plant-Wall) use preliminary object placements; GO-Renderer
on ScanNet++ is evaluated on a subset of the views.}
\label{tab:main}
\end{table*}
\begin{figure*}[t]
  \centering
  \gsetup{\linewidth}{480}{360}{4}{3868}%
  \ggrid{FIPT bedroom,FIPT bathroom,NFINITE}{Input,GS-IR,EAG-PT,GaSLight,MV-CoLight,LumiTokens,Ours,GT}{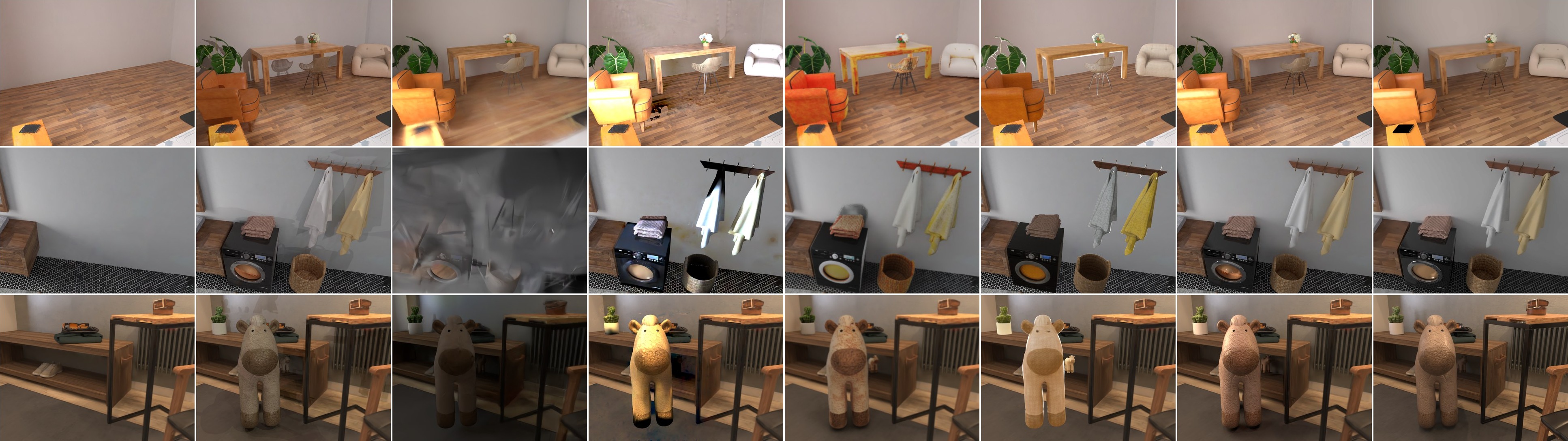}
  \caption{\textbf{Qualitative comparison on synthetic scenes}, cropped around the inserted
  objects. First column: input; last: ground truth (denoised for display). Ours reproduces object
  brightness and contact shadows; the baselines miss shadows or misjudge brightness.}
  \label{fig:qual}
\end{figure*}
\subsection{Emitter Radiance Estimation}
\label{sec:radiance}

\paragraph{Basis renders.}
With geometry, materials, and emitter placement fixed, the
remaining unknowns are the non-negative emitter radiances
$\bL=(L_1,\ldots,L_J)^\top\geq 0$. Since emitted radiance
enters the rendering equation linearly and the scene
transport is linear in the emitter radiances, the linear
image can be expressed as
\begin{equation}
\bI(\bL)=\sum_{j=1}^{J}L_j\bB_j,
\label{eq:linearity}
\end{equation}
where $\bB_j$ is the path-traced image obtained with emitter
$E_j$ at unit radiance and all other emitters disabled.
Each basis image includes the direct and indirect transport
originating from that emitter. To prevent the reconstructed appearance of an emitter from
being counted alongside its analytic representation, we remove
the bright Gaussians within a thin slab around $E_j$ when
rendering its own basis. In all other basis renders, these
Gaussians remain in the scene and act as reconstructed
surfaces. This treatment lets each basis represent a distinct
emitter hypothesis.

\paragraph{Clipped and unclipped observations.}
For each fitting view, let $\by$ denote the observed image
and let $\tau=0.98$ be the clipping threshold. Since a
saturated colour channel provides only a lower bound on the
underlying sensor response, we distinguish unclipped and
clipped observations per channel:
\[
\mathcal{U}_c=\{p:y_{p,c}<\tau\},
\qquad
\mathcal{C}_c=\{p:y_{p,c}\geq\tau\}.
\]

\paragraph{Objective.}
Let $g$ denote the sRGB encoding function or the fitted EMoR
response, and let $\bB_{p,c}\in\mathbb{R}^{J}$ collect the
unit-radiance basis values at pixel $p$ and colour channel
$c$. We estimate the emitter radiances independently per
channel by solving
\begin{equation}
\begin{aligned}
\min_{\bL_c\geq 0}\quad
&
\sum_{p\in\mathcal{U}_c}
\left[
g(\bB_{p,c}^{\top}\bL_c)-y_{p,c}
\right]^2
\\
&+
\sum_{p\in\mathcal{C}_c}
\left[
\tau-g(\bB_{p,c}^{\top}\bL_c)
\right]_+^2,
\end{aligned}
\label{eq:objective}
\end{equation}
where $[x]_+=\max(x,0)$. Unclipped observations are matched
directly, whereas clipped observations impose a one-sided
penalty that discourages predictions below the clipping
threshold without penalizing predictions above it.
This treatment matters because the brightest pixels are
precisely those that observe the emitters. Fitting them as
equalities would pull the radiances down towards the sensor
ceiling, and discarding them would lose the information that
the emitters are at least that bright. With the one-sided
term, the clipped pixels only bound the solution from below,
and its magnitude is set by the unsaturated walls, floor and
furniture, whose brightness depends on each emitter through
the basis images.

The problem contains $J$ unknowns per channel, typically
fewer than twenty, but can involve tens of millions of
residuals across all fitting views. We evaluate residuals
and Jacobians on the GPU and solve the resulting
$J\times J$ normal equations using Levenberg--Marquardt.
In linear space, the objective reduces to non-negative
least squares; the nonlinear camera response makes the
optimization nonlinear while keeping its dimensionality
low. In our experiments, the solution is insensitive to
initialization.

\paragraph{Suppressing spurious emitter candidates.}
Pixels directly observing a window or lamp may be saturated
and therefore provide only lower-bound constraints on the
predicted sensor response. Such observations alone cannot
reliably distinguish an emitter from a bright reflector.
Instead, the estimated radiances are constrained primarily
by how well each candidate explains the illumination of
unsaturated regions in the room. Consequently, a candidate
that contributes little to the observed room illumination
can receive near-zero radiance even if it was included in
the initial user annotations.

\subsection{Differential Compositing}
\label{sec:composite}

Replacing the original 3DGS rendering with a path-traced
image can discard view-dependent appearance and reconstruction
details. We therefore transfer the lighting changes induced
by the inserted object while retaining the original scene
appearance. We trace the scene from the same camera with and without the
inserted objects, obtaining $\bI_{\mathrm{with}}$ and
$\bI_{\mathrm{without}}$ in linear radiance. Given the object
mask $M$ and the original 3DGS rendering $\bI_{\mathrm{3dgs}}$,
we compute
\begin{equation}
\begin{aligned}
\bI_{\mathrm{out}}
={}&
M\,k\,\bI_{\mathrm{with}}
\\
&+
(1-M)\,\bI_{\mathrm{3dgs}}\odot
\operatorname{clip}\left(
\frac{\bI_{\mathrm{with}}}
{\bI_{\mathrm{without}}},
0,c
\right),
\end{aligned}
\label{eq:composite}
\end{equation}
where $\odot$ denotes element-wise multiplication,
$c=3$ limits the ratio, and $k$ is the median of
$\bI_{\mathrm{3dgs}}/\bI_{\mathrm{without}}$ over background
pixels to compensate for exposure differences.

The ratio captures the relative radiance change induced by
the inserted object: values below one correspond to
attenuation, such as cast shadows, while values above one
capture increased illumination, including indirect bounce
light. The original 3DGS rendering retains appearance details
that the path tracer does not model explicitly, and the
ratio modulates these details to incorporate the estimated
lighting changes. All compositing operations are performed
in linear radiance before applying the fitted camera response.
Optionally, we attenuate chromatic changes in the ratio using
\[
r\mapsto\bar r\left(\frac{r}{\bar r}\right)^\eta,
\]
where $\bar r$ is the channel mean and $\eta$ controls the
strength of the transferred bounce-light colour.

%% file: sec/4_experiments.tex
\section{Experiments}
\label{sec:exp}

\begin{figure*}[t]
  \centering
  \gsetup{0.75\linewidth}{480}{360}{4}{2900}%
  \ggrid{FIPT-Real,ScanNet++,Generated}{Input,GS-IR,GaSLight,MV-CoLight,LumiTokens,Ours}{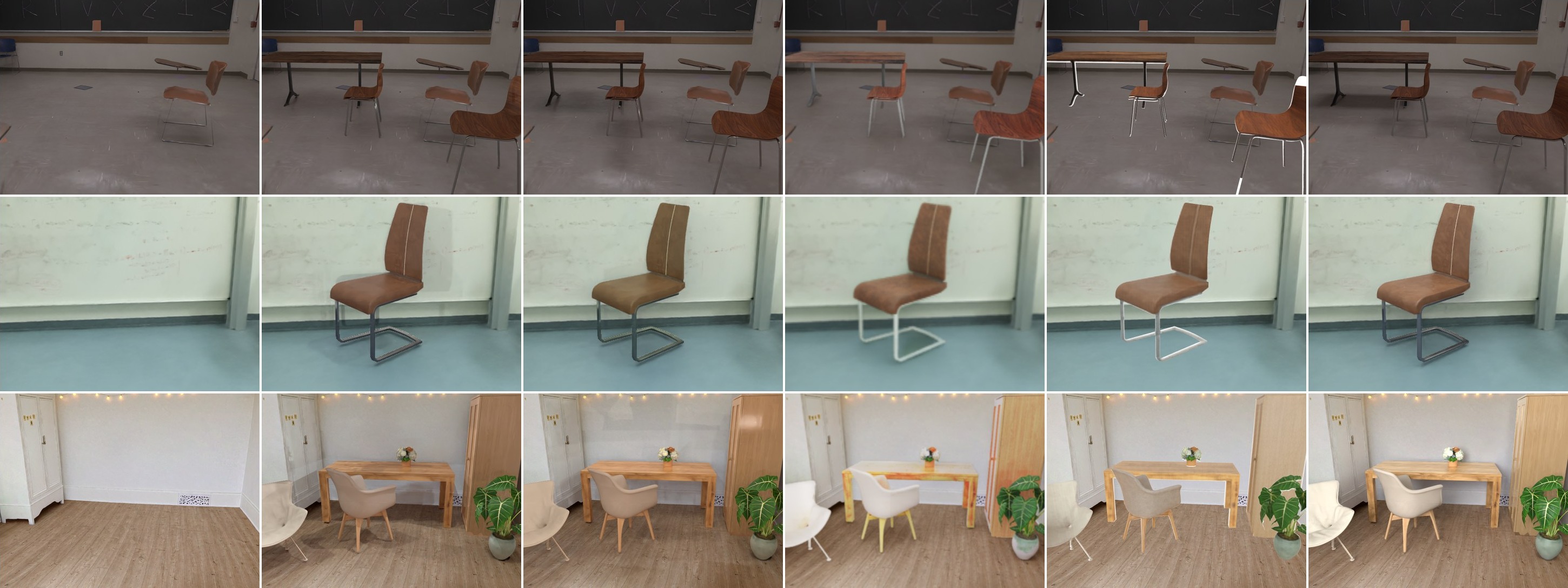}
  \caption{\textbf{Qualitative comparison on real and generated scenes} (columns as in \cref{fig:qual}
  without ground truth and EAG-PT, which needs linear HDR inputs).}
  \label{fig:qual_real}
\end{figure*}

\begin{figure}[t]
  \centering
  \gsetup{0.86\linewidth}{320}{160}{4}{968}%
  \ggrid{GI-GS,GS-IR,EAG-PT,Ours,GT}{Albedo,Roughness,Metallic}{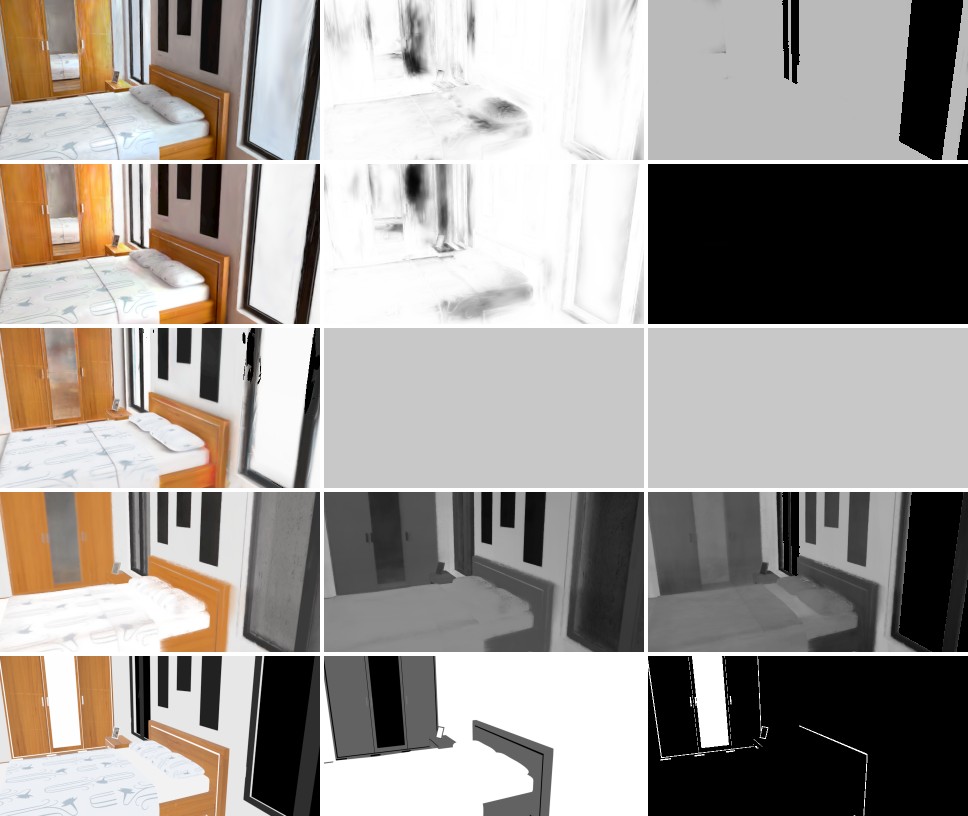}\\[3pt]
  \footnotesize\setlength{\tabcolsep}{3pt}
  \begin{tabular}{@{}lccccc@{}}
  \toprule
  & \multicolumn{3}{c}{Albedo} & Rough & Metal\\
  \cmidrule(lr){2-4}\cmidrule(lr){5-5}\cmidrule(lr){6-6}
  Method & PSNR$\uparrow$ & SSIM$\uparrow$ & LPIPS$\downarrow$ & L2$\downarrow$ & L2$\downarrow$\\
  \midrule
  GI-GS  & 12.44 & 0.642 & 0.418 & 0.142 & 0.515\\
  GS-IR  & 11.87 & 0.615 & 0.463 & \textbf{0.140} & \textbf{0.033}\\
  EAG-PT & \textbf{20.17} & 0.786 & 0.258 & -- & --\\
  \midrule
  Ours   & 19.93 & \textbf{0.814} & \textbf{0.218} & 0.155 & 0.161\\
  \bottomrule
  \end{tabular}
  \caption{\textbf{Recovered materials} on the FIPT bedroom (ours: RGB$\leftrightarrow$X prior) and
  scores averaged over the four FIPT rooms (albedo scale-aligned). Grey tiles, --:
  not produced. GS-IR's all-zero metallic scores well as only $1$--$6\%$ of the pixels are metal.}
  \label{fig:mat}
\end{figure}

\paragraph{Setup.}
We evaluate on synthetic scenes with path-traced ground truth (four FIPT~\cite{wu2023factorized}
rooms and two NFINITE~\cite{nfinite} scenes), real captures (two FIPT-Real~\cite{wu2023factorized}
and five ScanNet++~\cite{yeshwanth2023scannet++} scenes) and five 3DGS scenes generated on the
SpAItial platform. Objects, placements and lights are identical for every method, with three views
per scene. With ground truth we report $\Delta$PSNR, mask PSNR, shadow IoU and shadow RMSE, which
isolate the change caused by the inserted objects; without it, the no-reference VQAScore ratio
of LightHarmony3D~\cite{huang2026lightharmony3d}.
$\Delta$PSNR compares the change between the render with and without the objects against the
corresponding change in the ground truth, mask PSNR scores the object pixels, and the shadow IoU
and RMSE score the cast shadow region and its strength. Full-image
PSNR, SSIM and LPIPS are dominated by the background, which most methods share.

\paragraph{Implementation details.}
All our steps run on a single RTX~3090. For the FIPT bedroom, reconstruction takes $22$\,min,
the DiffusionRenderer prior $1.5$\,h per seed, back-projection $6$\,min and the radiance fit
$13.5$\,min. Inserting objects into one $1024{\times}576$ view takes $1.6$\,min: two path-traced
renders at $256$\,spp (with and without the objects) and the differential composite. EAG-PT, which
also path traces the scene, needs $14.4$\,min per frame at its $1024$\,spp, about $12\times$ more
at equal samples.

\begin{table}[t]
\centering
\footnotesize
\setlength{\tabcolsep}{2.5pt}
\resizebox{\linewidth}{!}{%
\begin{tabular}{@{}lcccc@{}}
\toprule
Variant & $\Delta$PSNR$\uparrow$ & Mask$\uparrow$ & IoU$\uparrow$ & RMSE$\downarrow$\\
\midrule
Point lights & 26.5 & 20.5 & 0.51 & 0.12\\
Env.\ map (PT/3DGS) & 27.6/27.4 & 22.0/20.7 & 0.56/0.55 & 0.11/0.12\\
\midrule
Bounces 1/2 & 25.7/30.8 & 14.5/24.9 & 0.25/0.63 & 0.33/0.12\\
Bounces 3/4 & 32.9/34.0 & 28.4/30.2 & 0.68/0.71 & 0.10/0.09\\
\midrule
GT window geometry & 35.4 & 31.7 & 0.80 & 0.08\\
Size $\times$0.8/1.25/1.5 & 34.2/35.4/35.4 & 31.1/31.4/31.3 & 0.80 & 0.08\\
Shift 5/10/20\,cm & 35.4/35.1/34.3 & 31.6/31.4/30.7 & 0.80 & 0.08\\
Shift 10\,cm (normal) & 35.3 & 30.4 & 0.80 & 0.08\\
\midrule
Shadow only/one-way & 31.3/27.5 & 31.4/30.1 & 0.58/0.00 & 0.12/0.36\\
Emissive 3DGS & 27.5 & 30.2 & 0.00 & 0.36\\
\midrule
Gray albedo & 23.5 & 15.6 & 0.61 & 0.17\\
w/o diff.\ compositing & 24.8 & 31.6 & 0.49 & 0.14\\
\midrule
\textbf{Ours (full)} & \textbf{35.4} & \textbf{31.6} & \textbf{0.80} & \textbf{0.08}\\
\bottomrule
\end{tabular}%
}
\caption{\textbf{Ablations} on the bedroom. Shifts are in the window plane unless marked.}
\label{tab:ablation_summary}
\end{table}

\subsection{Comparison}
\label{sec:comparison}
We compare our method with recent approaches in inverse rendering, illumination estimation, image harmonization, and feed-forward relighting. We use the released implementations of the baselines and follow their respective optimization protocols on our datasets. As shown in \Cref{tab:main}, our method achieves the best overall performance on FIPT and NFINITE, particularly in shadow reconstruction. Compared with inverse rendering methods, our near-field emitter representation yields more accurate shadows under a controlled rendering setup. Illumination estimation methods such as GaSLight struggle to recover accurate insertion shadows, while image harmonization methods such as MV-CoLight cannot reliably reproduce cast shadows through appearance adjustment alone. Similarly, feed-forward relighting methods produce plausible relit appearances but fail to reconstruct spatially accurate shadows. VQAScore barely separates the methods on the real and generated scenes. Qualitative comparisons are shown in \cref{fig:qual,fig:qual_real}. On the synthetic scenes, ours
reproduces the object brightness and the contact shadows, whereas the baselines miss the shadows or
misjudge the brightness (\cref{fig:qual}).

\paragraph{Material recovery.}
\Cref{fig:mat} compares the recovered materials. GS-IR and GI-GS bake shading and window light
into the albedo of these texture-poor rooms, $6$--$8$\,dB below our aligned albedo
($19.9$\,dB on FIPT). EAG-PT reaches a similar albedo ($20.2$\,dB) but recovers diffuse albedo
only. Our prior also provides roughness and metallic, although the metallic channel spills onto
diffuse surfaces (\cref{fig:mat}).

\subsection{Ablation}
\label{sec:ablations}

\begin{figure}[t]
\centering
\gsetup{0.93\linewidth}{500}{313}{6}{1512}\gfixed{20pt}%
\newcommand{\abq}[3][]{\gstrip[r]{#2}{imgs/abl_qual_bedroom_f_r#3.jpg}\vskip\ggw}%
\vbox{\offinterlineskip
\ghead{Point lights,Env.\ map}\abq{Light}{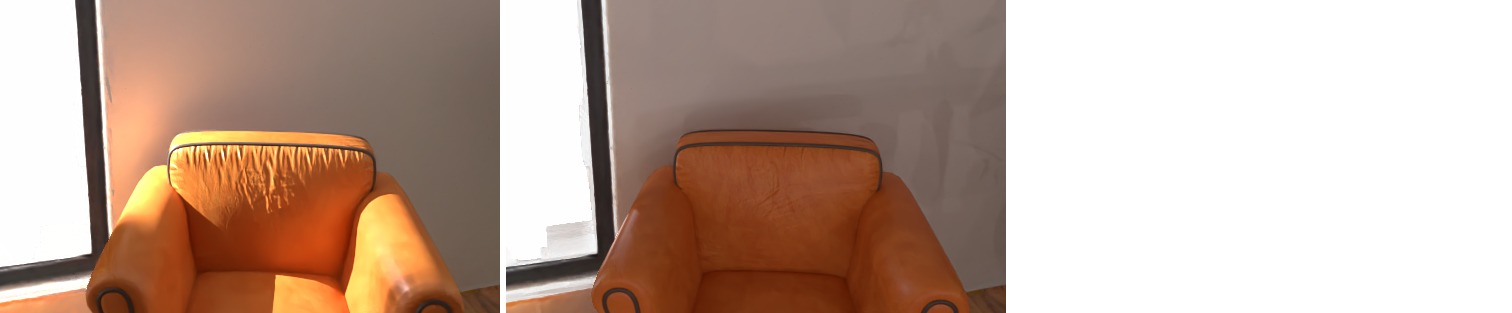}
\ghead{1 bounce,2 bounces}\abq{Transport}{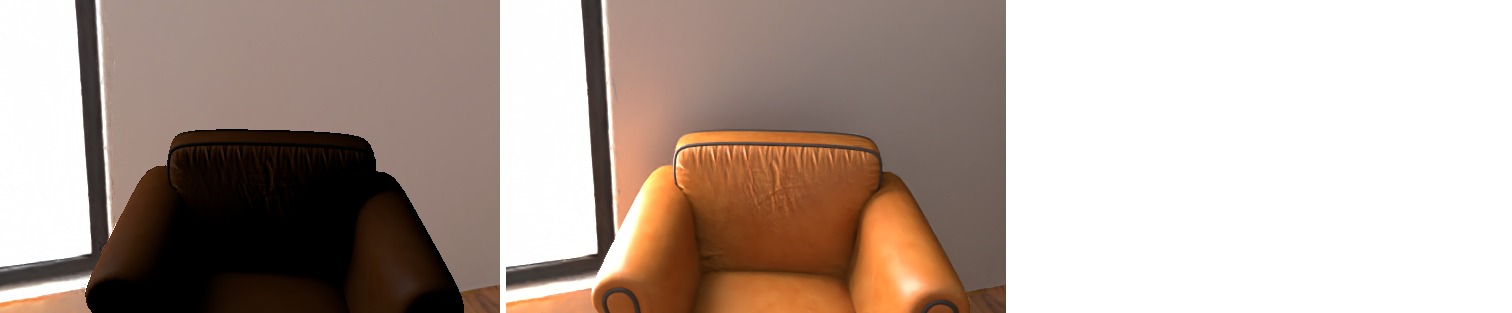}
\ghead{Shadow only,One-way,Emissive 3DGS}\abq{Interaction}{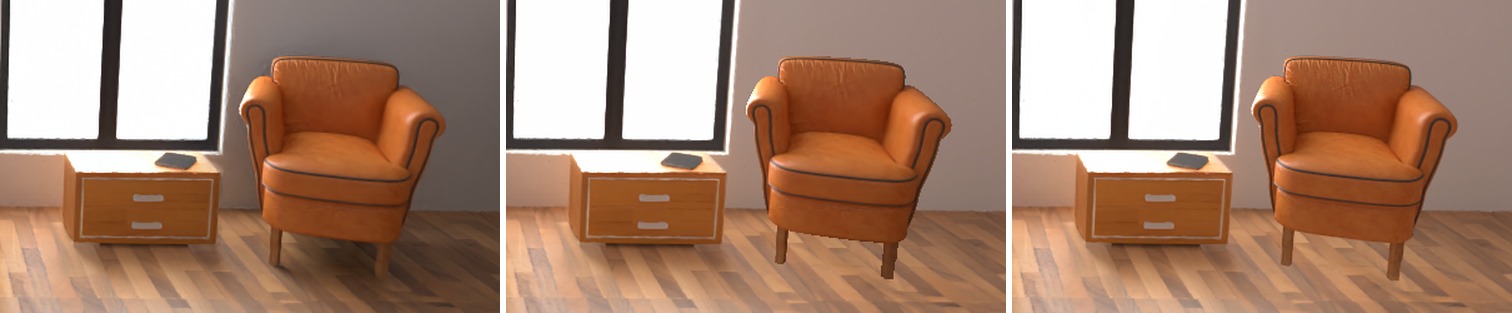}
\ghead{Gray albedo,\begin{tabular}[b]{@{}c@{}}Gray albedo\\(PT)\end{tabular},\begin{tabular}[b]{@{}c@{}}w/o diff.\\comp.\end{tabular}}\abq{Material, comp.}{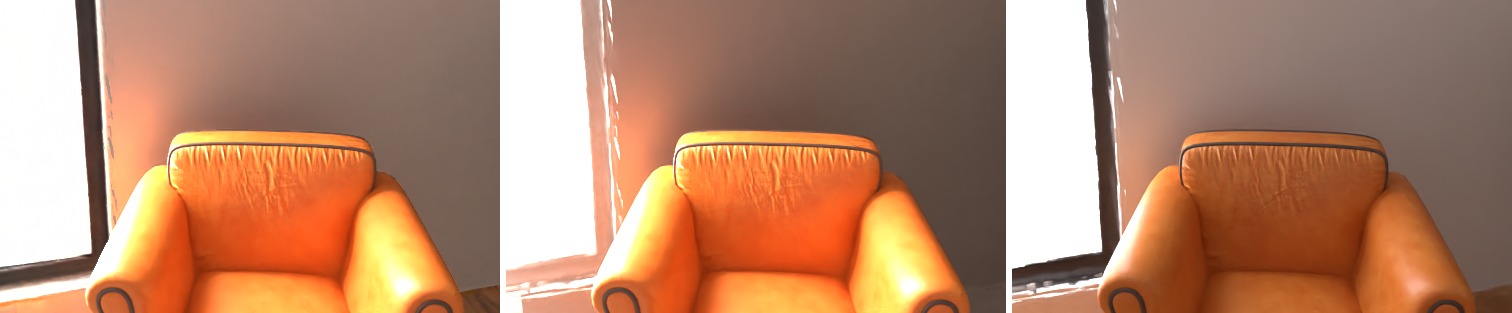}
\abq{\textbf{Ours}}{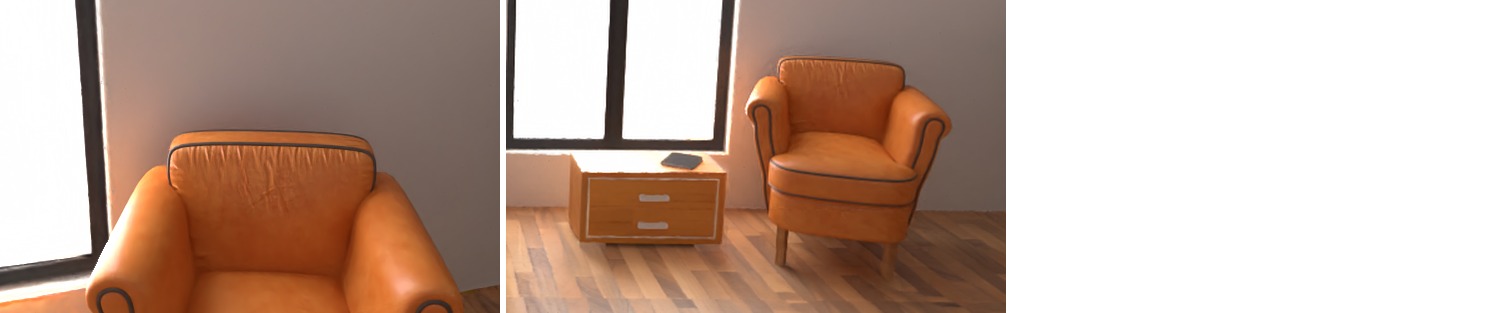}
\gstrip[r]{\textbf{GT}}{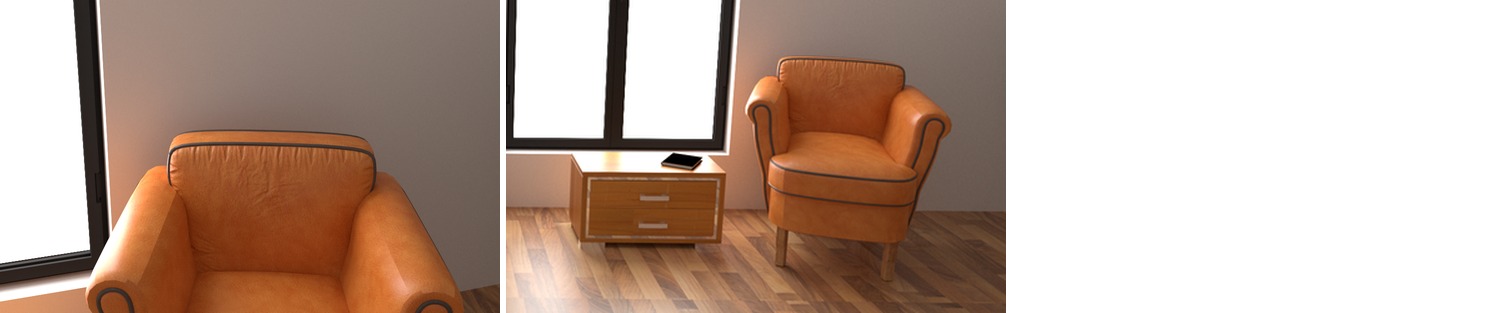}}
\caption{\textbf{Ablations, qualitative} (bedroom), one row per ablation; composites unless marked PT.
Point lights over-brighten, an environment map darkens; direct light is far too dark; without the
object-to-room bounce the wall misses reflected light; a gray albedo distorts the room (PT), compositing
hides it. Last rows: ours and GT, two views.}
\label{fig:abl_qual}
\end{figure}

\paragraph{Near-field area lights.}
Point lights and environment maps lose $8$-$9$\,dB $\Delta$PSNR and a third of the shadow IoU
(\cref{tab:ablation_summary}): points ignore a window's extent, an environment map its
falloff (\cref{fig:abl_qual}).

\paragraph{Multi-bounce transport.}
Direct light alone yields far too dark objects and shadows ($25.7$\,dB); every bounce helps,
and four bounces are within $1.4$\,dB of the full tracer.

\paragraph{Two-way interaction.}
Without the light bounced off the object into the room (shadow only, one-way, emissive 3DGS)
the object region stays plausible, but the shadow IoU falls to $0.58$ or $0$
(\cref{fig:abl_qual}).

\paragraph{Materials and compositing.}
A gray albedo instead of the material prior halves the mask PSNR ($15.6$\,dB), since both the
fitted lights and the shading depend on the albedo. Using our full path-traced frame instead of
the composite keeps the object but drops $\Delta$PSNR to $24.8$\,dB: the re-rendered room
inherits errors such as the missing mirror reflection, which differential compositing cancels.

\paragraph{Robustness to emitter geometry.}
Ground-truth windows, sizes $\times0.8$--$1.5$ and shifts up to $20$\,cm cost at most $1.3$\,dB
$\Delta$PSNR and keep the shadow IoU at $0.80$ (\cref{tab:ablation_summary}).

\subsection{Limitations}

Our method has two main limitations. First, the light sources must be marked
manually: the user selects windows and lamps with SAM in one or a few views,
although only approximate masks are needed. Second, it relies on prior
information: albedo, roughness, metallic and normals come from 2D estimators
(RGB$\leftrightarrow$X or DiffusionRenderer), so errors in these priors carry
over to the fitted lights and the insertion.

%% file: sec/5_conclusion.tex
\section{Conclusion}
\label{sec:conclusion}

We recover sampleable area emitters (geometry, orientation and absolute
radiance) from ordinary LDR multi-view captures of indoor scenes by treating
saturated pixels as censored observations. With a stochastic-transparency path
tracer over the Gaussians, inserted objects receive spatially varying shading,
contact shadows and bounce light without degrading the reconstruction. With geometry and
materials fixed, the light estimate is a low-dimensional fit
over path-traced basis images. We also show that full-image metrics barely reflect lighting,
and give a protocol that isolates the insertion effect.